\documentclass[10pt,twocolumn,letterpaper]{article}

\usepackage{cvpr}              

\usepackage{graphicx}
\usepackage{amsmath}
\usepackage{amssymb}
\usepackage{pifont}
\usepackage{booktabs}
\usepackage[table,xcdraw]{xcolor}
\usepackage{makecell}
\usepackage{listings}
\usepackage{multirow}

\usepackage[pagebackref,breaklinks,colorlinks]{hyperref}

\usepackage[capitalize]{cleveref}
\crefname{section}{Sec.}{Secs.}
\Crefname{section}{Section}{Sections}
\Crefname{table}{Table}{Tables}
\crefname{table}{Tab.}{Tabs.}

\newcommand{\model}{\mbox{\textsc{{VisProg}}\xspace}}
\newcommand{\gpt}{\mbox{\textsc{{GPT-3}}\xspace}}
\newcommand{\gqa}{\mbox{\textsc{{GQA}}\xspace}}
\newcommand{\vqavtwo}{\mbox{\textsc{{VQAv2}}\xspace}}
\newcommand{\nlvr}{\mbox{\textsc{{NLVRv2}}\xspace}}
\newcommand{\vilt}{\mbox{\textsc{{ViLT}}\xspace}}
\newcommand{\viltvqa}{\mbox{\textsc{{ViLT-vqa}}\xspace}}
\newcommand{\viltnlvr}{\mbox{\textsc{{ViLT-nlvr}}\xspace}}
\newcommand{\clip}{\mbox{\textsc{{CLIP}}\xspace}}
\newcommand{\reinf}{\mbox{\textsc{{Reinforce}}\xspace}}
\newcommand{\cmark}{\ding{51}}%
\newcommand{\xmark}{\ding{55}}%

\definecolor{codegreen}{rgb}{0,0.6,0}
\definecolor{codegray}{rgb}{0.5,0.5,0.5}
\definecolor{codepurple}{rgb}{0.58,0,0.82}
\definecolor{backcolour}{rgb}{0.98,0.98,0.97}

\lstdefinestyle{mystyle}{
    backgroundcolor=\color{backcolour},   
    commentstyle=\color{codegreen},
    keywordstyle=\color{magenta},
    numberstyle=\tiny\color{codegray},
    stringstyle=\color{codepurple},
    basicstyle=\ttfamily\footnotesize,
    breakatwhitespace=false,         
    breaklines=true,                 
    captionpos=b,                    
    keepspaces=true,                 
    numbersep=5pt,                  
    showspaces=false,                
    showstringspaces=false,
    showtabs=false,                  
    tabsize=2,
    frame=single,
}

\def\cvprPaperID{2891} 
\def\confName{CVPR}
\def\confYear{2023}

\begin{document}

\title{Visual Programming: Compositional visual reasoning without training}

\author{Tanmay Gupta, Aniruddha Kembhavi\\
PRIOR @ Allen Institute for AI\\
\url{https://prior.allenai.org/projects/visprog}
}

\twocolumn[{%
\renewcommand\twocolumn[1][]{#1}%
\maketitle
\begin{center}
    \centering
    \captionsetup{type=figure}
    \includegraphics[width=\textwidth]{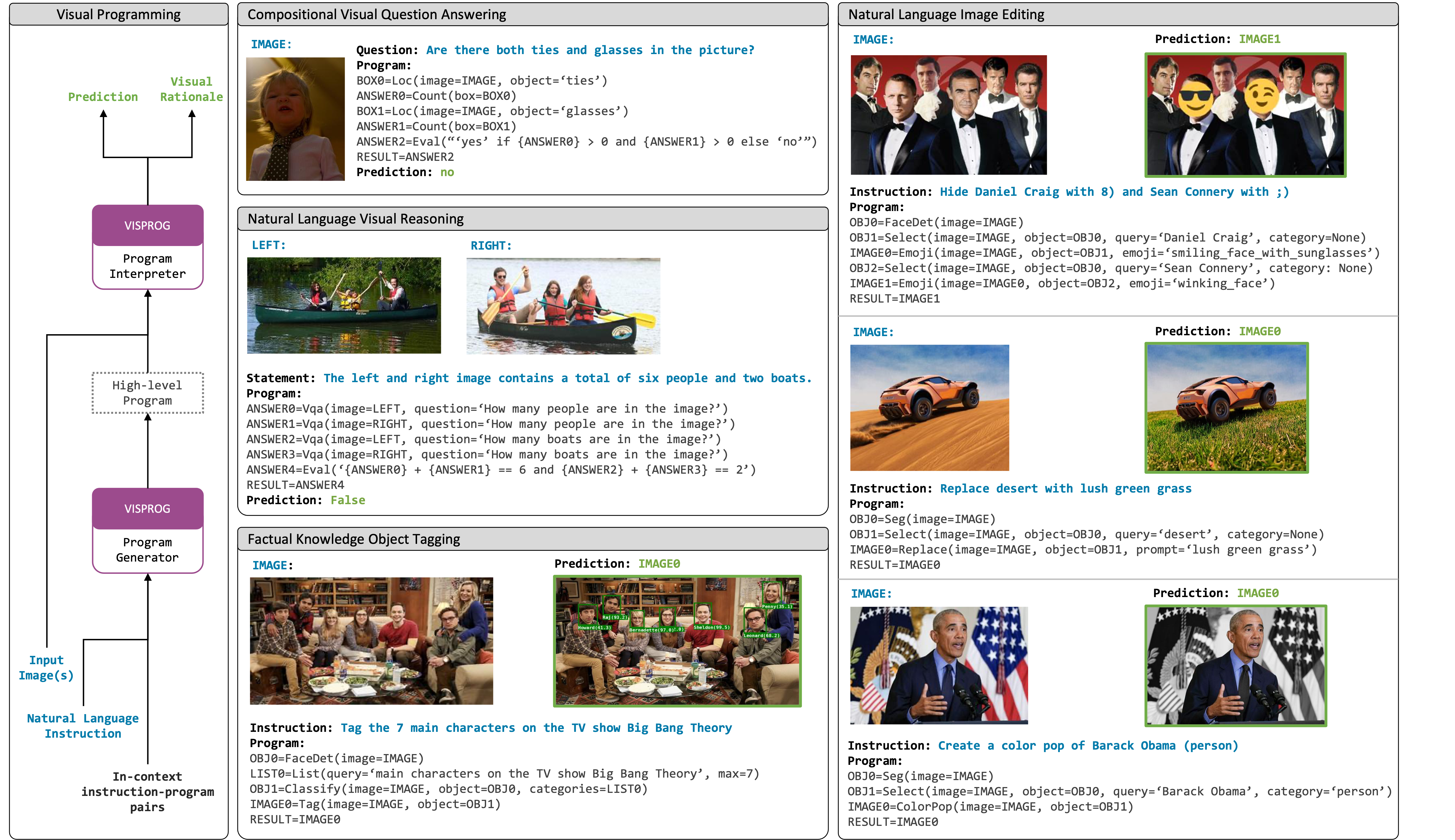}
    \captionof{figure}{\textbf{\model\ is a modular and interpretable neuro-symbolic system for compositional visual reasoning.} Given a few examples of natural language instructions and the desired high-level programs, \model\ generates a program for any new instruction using \textit{in-context learning} in \gpt\ and then executes the program on the input image(s) to obtain the prediction. \model\ also summarizes the intermediate outputs into an interpretable \textit{visual rationale} (Fig.~\ref{fig:rationale}). We demonstrate \model\ on tasks that require composing a diverse set of modules for image understanding and manipulation, knowledge retrieval, and arithmetic and logical operations.}
    \label{fig:teaser}
\end{center}%
}]

\begin{abstract}

We present \model, a neuro-symbolic approach to solving complex and compositional visual tasks given natural language instructions. \model\ avoids the need for any task-specific training. Instead, it uses the in-context learning ability of large language models to generate python-like modular programs, which are then executed to get both the solution and a comprehensive and interpretable rationale. Each line of the generated program may invoke one of several off-the-shelf computer vision models, image processing subroutines, or python functions to produce intermediate outputs that may be consumed by subsequent parts of the program. We demonstrate the flexibility of \model\ on 4 diverse tasks - compositional visual question answering, zero-shot reasoning on image pairs, factual knowledge object tagging, and language-guided image editing. We believe neuro-symbolic approaches like \model\ are an exciting avenue to easily and effectively expand the scope of AI systems to serve the long tail of complex tasks that people may wish to perform.

\end{abstract}

\section{Introduction}
\label{sec:intro}

The pursuit of general purpose AI systems has lead to the development of capable end-to-end trainable models\cite{Alayrac2022Flamingo,Lu2022UnifiedIOAU,Gupta2021TowardsGP,Kamath2022WeblySC,t5,gpt3,Reed2022Gato}, many of which aspire to provide a simple natural language interface for a user to interact with the model. The predominant approach to building these systems has been massive-scale unsupervised pretraining followed by supervised multitask training. However, this approach requires a well curated dataset for each task that makes it challenging to scale to the infinitely long tail of complex tasks we would eventually like these systems to perform. In this work, we explore the use of large language models to tackle the long tail of complex tasks by decomposing these tasks described in natural language into simpler steps that may be handled by specialized end-to-end trained models or other programs.

 
Imagine instructing a vision system to ``Tag the 7 main characters on the TV show Big Bang Theory in this image." To perform this task, the system first needs to understand the intent of the instruction and then perform a sequence of steps - detect the faces, retrieve list of main characters on Big Bang Theory from a knowledge base, classify faces using the list of characters, and tag the image with recognized character's faces and names. While different vision and language systems exist to perform each of these steps, executing this task described in natural language is beyond the scope of end-to-end trained systems. 


We introduce \model\ which inputs visual data (a single image or a set of images) along with a natural language instruction, generates a sequence of steps, a \textit{visual program} if you will, and then executes these steps to produce the desired output. Each line in a visual program invokes one among a wide range of modules currently supported by the system. Modules may be off-the-shelf computer vision models, language models, image processing subroutines in OpenCV~\cite{bradski2000opencv}, or arithmetic and logical operators. Modules consume inputs that are produced by executing previous lines of code and output intermediate results that can be consumed downstream. In the example above, the visual program generated by \model\ invokes a face detector~\cite{Li2019DSFD}, \gpt\ \cite{gpt3} as a knowledge retrieval system, and \clip\ \cite{clip} as an open-vocabulary image classifier to produce the desired output (see Fig.~\ref{fig:teaser}).


\begin{figure}[t]
  \centering
  \includegraphics[width=1.0\linewidth]{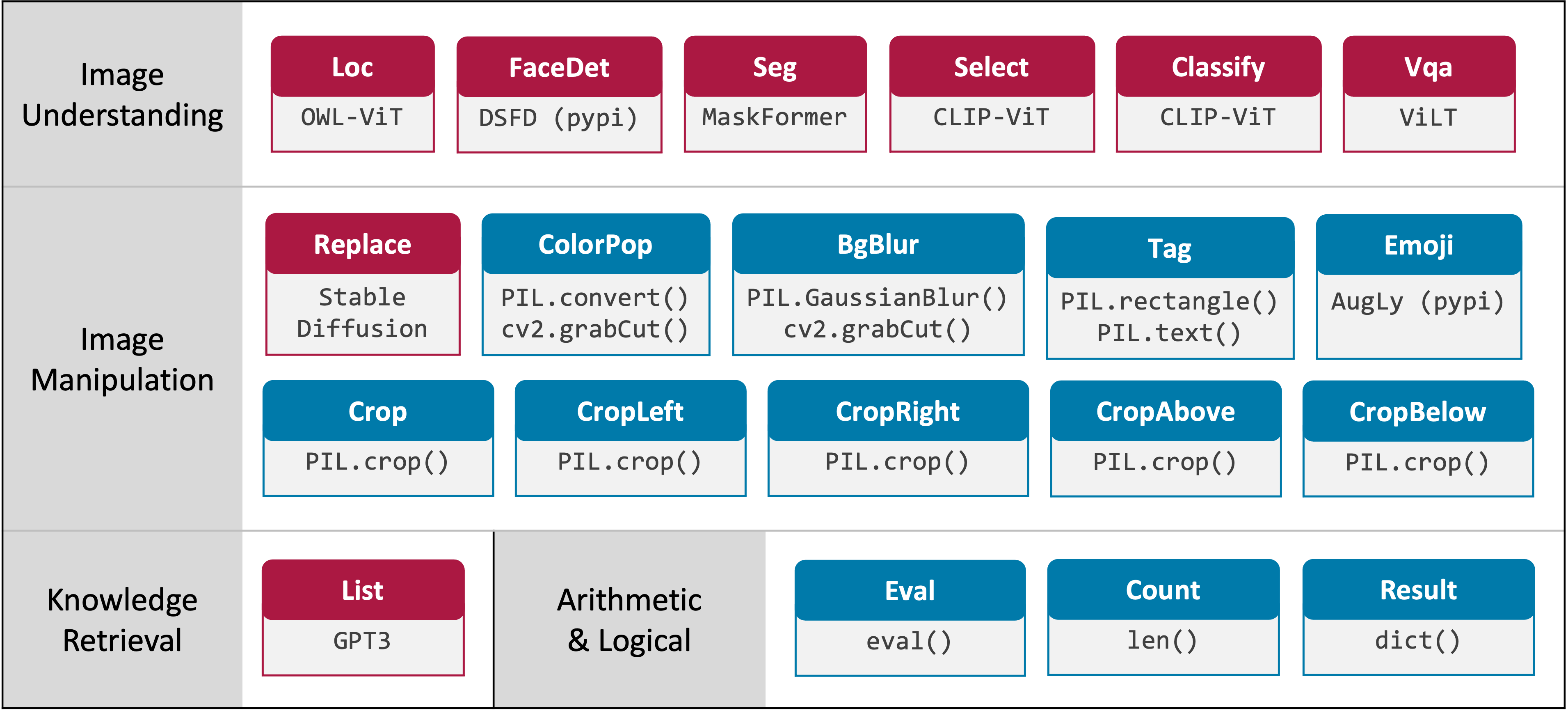} 
  \vspace{-1.5em}
   \caption{\textbf{Modules currently supported in \model.} Red modules use neural models (OWL-ViT~\cite{Minderer2022owlvit}, DSFD~\cite{Li2019DSFD}, MaskFormer~\cite{Cheng2021Maskformer}, CLIP~\cite{clip}, ViLT~\cite{Kim2021ViLT}, and Stable Diffusion~\cite{Rombach2022stableDiffusion}). Blue modules use image processing and other python subroutines. These modules are invoked in programs generated from natural language instructions. Adding new modules to extend \model's capabilities is straightforward (Code.~\ref{lst:module}).}
   \label{fig:modules}
   \vspace{-1.5em}
\end{figure}

\model\ improves upon previous methods for generating and executing programs for vision applications. For the visual question answering (VQA) task, Neural Module Networks (NMN) \cite{Andreas2016NeuralMN,Hu2017N2NMN,Johnson2017InferringAE,Hu2018StackNMN} compose a question-specific, end-to-end trainable network from specialized, differentiable neural modules. These approaches either use brittle, off-the-shelf semantic parsers to deterministically compute the layout of modules, or learn a layout generator through weak answer supervision via \reinf\ \cite{williams1992reinforce}. In contrast, \model\ uses a powerful language model (\gpt) and a small number of in-context examples to create complex programs \textit{without} requiring any training\footnote{We use ``training" to refer to gradient-based learning to differentiate it from in-context learning which only involves a feedforward pass.}. Programs created by \model\ also use a higher-level of abstraction than NMNs and invoke trained state-of-the-art models and non-neural python subroutines (Fig.~\ref{fig:modules}). These advantages make \model\ an easy-to-use, performant, and modular neuro-symbolic system.

\model\ is also highly interpretable. First, \model\ produces easy-to-understand programs which a user can verify for logical correctness. Second, by breaking down the prediction into simple steps, \model\ allows a user to inspect the outputs of intermediate steps to diagnose errors and if required, intervene in the reasoning process. Altogether, an executed program with intermediate step results (\eg text, bounding boxes, segmentation masks, generated images, etc.) linked together to depict the flow of information serves as a \textit{visual rationale} for the prediction.  

To demonstrate its flexibility, we use \model\ for 4 different tasks that share some common skills (\eg for image parsing) while also requiring some degree of specialized reasoning and visual manipulation capabilities. These tasks are - (i) compositional visual question answering; (ii) zero-shot natural language visual reasoning (NLVR) on image pairs; (iii) factual knowledge object tagging from natural language instructions; and (iv) language-guided image editing. We emphasize that neither the language model nor any of the modules are finetuned in any way. Adapting \model\ to any task is as simple as providing a few in-context examples consisting of natural language instructions and the corresponding programs. While easy to use, \model\ shows an impressive gain of $2.7$ points over a base VQA model on the compositional VQA task, strong zero-shot accuracy of $62.4\%$ on NLVR without ever training on image pairs, and delightful qualitative and quantitative results on knowledge tagging and image editing tasks.


Our key contributions include - (i) \model\ - a system that uses the in-context learning ability of a language model to generate visual programs from natural language instructions for compositional visual tasks (Sec.~\ref{sec:visprog}); (ii) demonstrating the flexibility of \model\ on complex visual tasks such as factual knowledge object tagging and language guided image editing (Secs.~\ref{sec:knowtag} and~\ref{sec:imgedit}) that have eluded or seen limited success with a single end-to-end model; and (iii) producing \textit{visual rationales} for these tasks and showing their utility for error analysis and user-driven instruction tuning to improve \model's performance significantly (Sec.~\ref{sec:utility}). 





\section{Related Work}
\label{sec:related_work}
Neuro-symbolic approaches have seen renewed momentum owing to the incredible understanding, generation, and in-context learning capabilities of large language models (LLMs). We now discuss previous program generation and execution approaches for visual tasks, recent work in using LLMs for vision, and advances in reasoning methods for language tasks. \\

\noindent\textbf{Program generation and execution for visual tasks.} Neural module networks (NMN)~\cite{Andreas2016NeuralMN} pioneered modular and compositional approaches for the visual question answering (VQA) task. NMNs compose neural modules into an end-to-end differentiable network. While early attempts use off-the-shelf parsers~\cite{Andreas2016NeuralMN}, recent methods~\cite{Hu2017N2NMN,Johnson2017InferringAE,Hu2018StackNMN} 
learn the layout generation model jointly with the neural modules using \reinf\ \cite{williams1992reinforce} and weak answer supervision. 
While similar in spirit to NMNs, \model\ has several advantages over NMNs. First, \model\ generates \textit{high-level} programs that invoke trained state-of-the-art neural models and other python functions at intermediate steps as opposed to generating end-to-end neural networks. This makes it easy to incorporate symbolic, non-differentiable modules. Second, \model\ leverages the \textit{in-context learning} ability of LLMs~\cite{gpt3} to generate programs by prompting the LLM (\gpt) with a natural language instruction (or a visual question or a statement to be verified) along with a few examples of similar instructions and their corresponding programs thereby removing the need to train specialized program generators for each task. \\ 

\noindent\textbf{LLMs for visual tasks.} LLMs and in-context learning have been applied to visual tasks. PICa~\cite{Yang2022pica} uses LLMs for a knowledge-based VQA~\cite{Marino2019OKVQA} task. PICa represents the visual information in images as text via captions, objects, and attributes and feeds this textual representation to \gpt\ along with the question and in-context examples to directly generate the answer. Socratic models (SMs)~\cite{zeng2022socraticmodels}, compose pretrained models from different modalities such as language (BERT~\cite{Devlin2019BERT}, GPT-2~\cite{Radford2019gpt2}), vision-language (\clip\ \cite{clip}), and audio-language (mSLAM~\cite{Bapna2022mSLAM}), to perform a number of zero-shot tasks, including image captioning, video-to-text retrieval, and robot planning. However, in SMs the composition is pre-determined and fixed for \textit{each task}. In contrast, \model\ determines how to compose models for \textit{each instance} by generating programs based on the instruction, question, or statement. We demonstrate \model's ability to handle complex instructions that involve diverse capabilities (20 modules) and varied input (text, image, and image pairs), intermediate (text, image, bounding boxes, segmentation masks), and output modalities (text and images). Similar to \model, ProgPrompt~\cite{Singh2022ProgPrompt} is a concurrent work that demonstrates the ability of LLMs to generate python-like situated robot action plans from natural language instructions. While ProgPrompt modules (such as ``find" or ``grab") take strings (typically object names) as input, \model\ programs are more general. In each step in a \model\ program, a module could accept multiple arguments including strings, numbers, arithmetic and logical expressions, or arbitrary python objects (such as \lstinline{list()} or \lstinline{dict()} instances containing bounding boxes or segmentation masks) produced by previous steps. \\

\noindent\textbf{Reasoning via Prompting in NLP.} There is a growing body of literature~\cite{Khot2021TalkToAgent,Kojima2022ZSReason} on using LLMs for language reasoning tasks via prompting. Chain-of-Thought (CoT) prompting~\cite{Wei2022CoT}, where a language model is prompted with in-context examples of inputs, chain-of-thought rationales (a series of intermediate reasoning steps), and outputs, has shown impressive abilities for solving math reasoning problems. While CoT relies on the ability of LLMs to both generate a reasoning path and execute it, approaches similar to \model\ have been applied to language tasks, where a \textit{decomposer pompt}~\cite{Khot2022DecomposedPrompting} is used first to generate a sequence of sub-tasks which are then handled by sub-task handlers. 


\section{Visual Programming}
\label{sec:visprog}

\begin{figure}[t]
  \centering
  \includegraphics[width=1.0\linewidth]{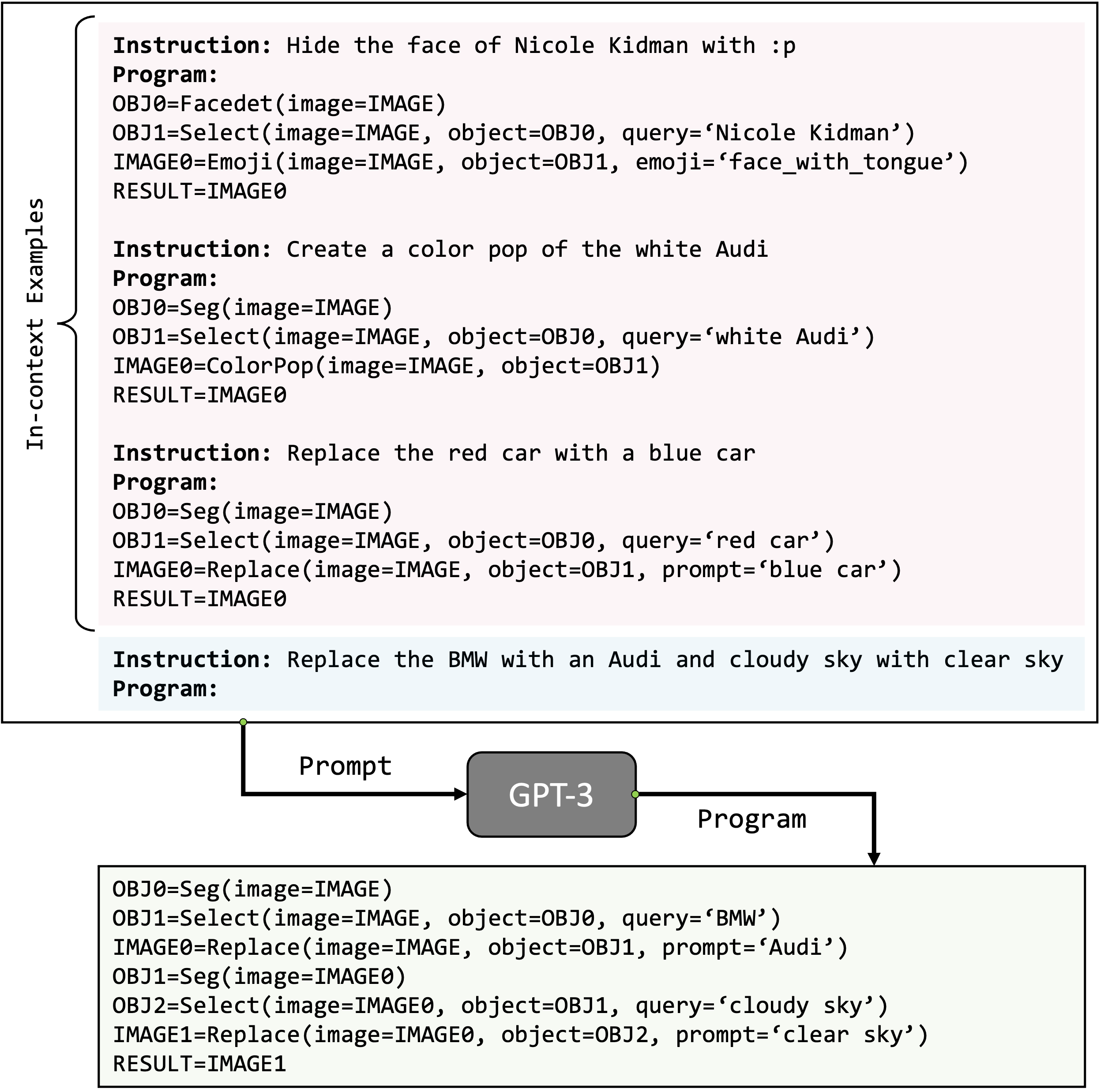}
   \caption{\textbf{Program generation in \model.}}
   \label{fig:prompt}
\end{figure}
Over the last few years, the AI community has produced high-performance, task-specific models for many vision and language tasks such as object detection, segmentation, VQA, captioning, and text-to-image generation. While each of these models solves a well-defined but narrow problem, the tasks we usually want to solve in the real world are often broader and loosely defined. 

To solve such practical tasks, one has to either collect a new task-specific dataset, which can be expensive, or meticulously compose a program that invokes multiple neural models, image processing subroutines (\eg image resizing, cropping, filtering, and colorspace conversions), and other computation (\eg database lookup, or arithmetic and logical operations). Manually creating these programs for the infinitely long tail of complex tasks we encounter daily not only requires programming expertise but is also slow, labor intensive, and ultimately insufficient to cover the space of all tasks. What if, we could describe the task in natural language and have an AI system generate and execute the corresponding visual program without any training? \\

\noindent\textbf{Large language models for visual programming.} Large language models such as \gpt\ have shown a remarkable ability to generalize to new samples for a task having seen a handful of input and output demonstrations \textit{in-context}. For example, prompting \gpt\ with two English-to-French translation examples and a new English phrase

\begin{lstlisting}[frame=none,backgroundcolor=\color{white},xleftmargin=.1\textwidth, xrightmargin=.1\textwidth]
good morning -> bonjour
good day -> bonne journée
good evening ->
\end{lstlisting}
produces the French translation ``bonsoir". Note that we did not have to finetune \gpt\ to perform the task of translation on the thrid phrase. \model\ uses this in-context learning ability of \gpt\ to output visual programs for natural language instructions.

Similar to English and French translation pairs in the example above, we prompt \gpt\ with pairs of instructions and the desired high-level program. Fig.~\ref{fig:prompt} shows such a prompt for an image editing task. The programs in the in-context examples are manually written and can typically be constructed without an accompanying image. Each line of a \model\ program, or a \textbf{program step}, consists of the name of a \textbf{module}, module's input argument names and their values, and an output variable name. \model\ programs often use output variables from past steps as inputs to future steps. We use descriptive module names (e.g. ``Select", ``ColorPop", ``Replace"), argument names (e.g. ``image", ``object", ``query"), and variable names (e.g. ``IMAGE", ``OBJ") to allow \gpt\ to understand the input and output type, and function of each module. During execution the output variables may be used to store arbitrary data types. For instance ``OBJ"s are list of objects in the image, with mask, bounding box, and text (\eg category name) associated with each object.

These in-context examples are fed into \gpt\ along with a new natural language instruction. Without observing the image or its content, \model\ generates a program (bottom of Fig.~\ref{fig:prompt}) that can be executed on the input image(s) to perform the described task. \\ 


\begin{lstlisting}[language=Python,basicstyle=\ttfamily\footnotesize,label={lst:module},caption=\textbf{Implementation of a \model\ module.},linewidth=1\columnwidth,xleftmargin=0\columnwidth]
class VisProgModule():
  def __init__(self):
    # load a trained model; move to GPU

  def html(self,inputs: List,output: Any):
    # return an html string visualizing step I/O
      
  def parse(self,step: str):
    # parse step and return list of input values 
    # and variables, and output variable name

  def execute(self,step: str,state: Dict):
    inputs, input_var_names, output_var_name = \
      self.parse(step)

    # get values of input variables from state
    for var_name in input_var_names:
      inputs.append(state[var_name])
    
    # perform computation using the loaded model
    output = some_computation(inputs)
    
    # update state
    state[output_var_name] = output
    
    # visual summary of the step computation
    step_html = self.html(inputs,output)
    return output, step_html
\end{lstlisting}

\noindent\textbf{Modules.} \model\ currently supports 20 modules (Fig.~\ref{fig:modules}) for enabling capabilities such as image understanding, image manipulation (including generation), knowledge retrieval, and performing arithmetic and logical operations. In \model, each module is implemented as a Python class (Code.~\ref{lst:module}) that has methods to: (i) \textit{parse} the line to extract the input argument names and values, and the output variable name; (ii) \textit{execute} the necessary computation that may involve trained neural models and update the program state with the output variable name and value; and (iii) summarize the step's computation visually using \textit{html} (used later to create a \textit{visual rationale}). Adding new modules to \model\ simply requires implementing and registering a module class, while the execution of the programs using this module is handled automatically by the \model\ interpreter, which is described next. \\

\begin{figure}[t]
  \centering
  \includegraphics[width=\linewidth]{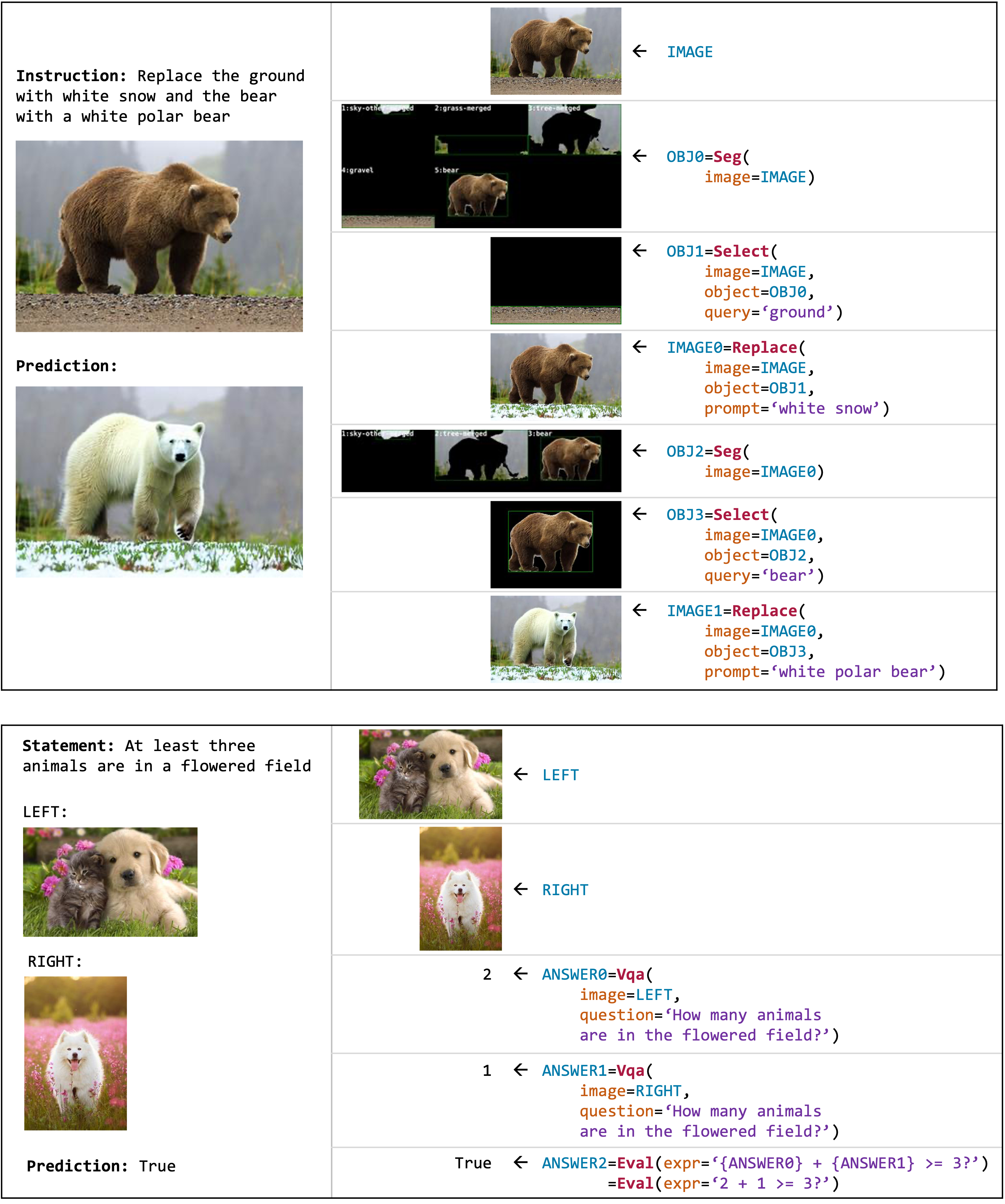}
   \caption{\textbf{Visual rationales generated by \model.} These rationales visually summarize the input and output of each computational step in the generated program during inference for an image editing (top) and NLVR task (bottom).} 
   \label{fig:rationale}
   \vspace{-1em}
\end{figure}

\noindent\textbf{Program Execution.} The program execution is handled by an \textbf{interpreter}. The interpreter initializes the program state (a dictionary mapping variables names to their values) with the inputs, and steps through the program line-by-line while invoking the correct module with the inputs specified in that line. After executing each step, the program state is updated with the name and value of the step's output. \\


\noindent\textbf{Visual Rationale.} In addition to performing the necessary computation, each module class also implements a method called \lstinline{html()} to visually summarize the inputs and outputs of the module in an HTML snippet. The interpreter simply stitches the HTML summary of all program steps into a visual rationale (Fig.~\ref{fig:rationale}) that can be used to analyze the logical correctness of the program as well as inspect the intermediate outputs. The visual rationales also enable users to understand reasons for failure and tweak the natural language instructions minimally to improve performance. See Sec.~\ref{sec:utility} for more details.

\section{Tasks}
\begin{figure}[t]
  \centering
  \includegraphics[width=1.0\linewidth]{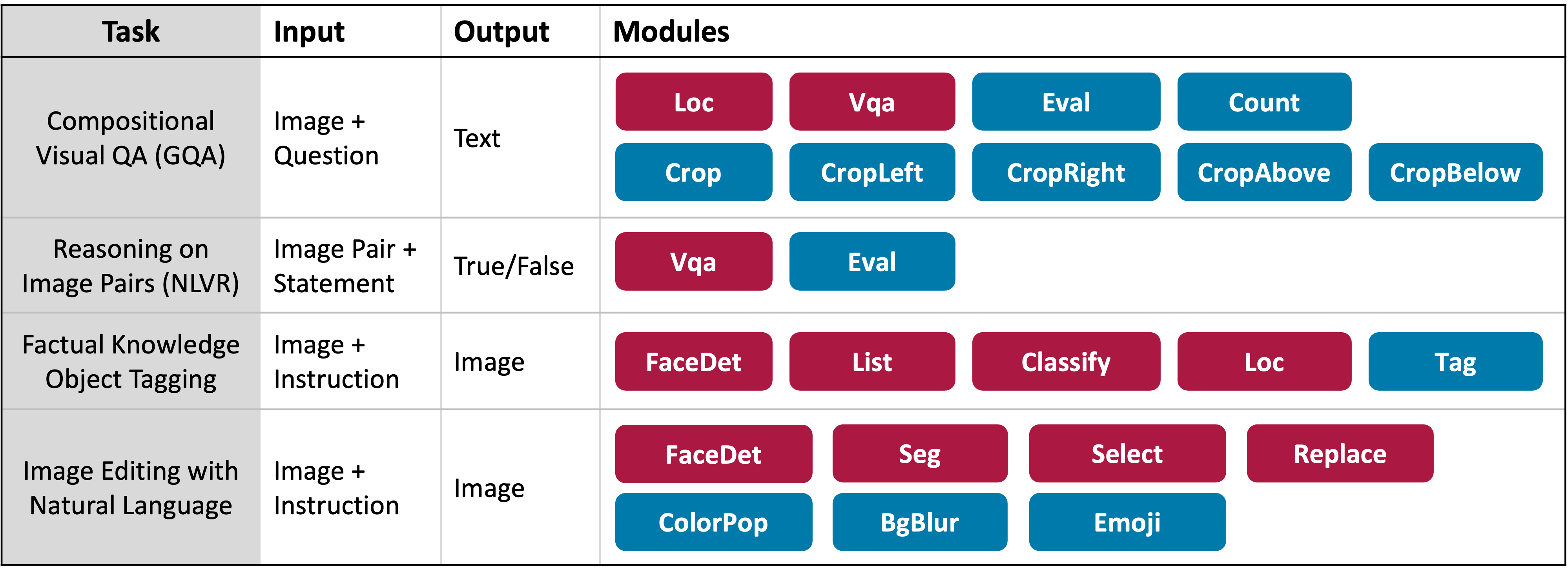}

   \caption{\textbf{We evaluate \model\ on a diverse set of tasks.} The tasks span a variety of inputs and outputs and reuse modules (Loc, FaceDet, VQA) whenever possible.}
   \label{fig:task_io_modules}
   \vspace{-1em}
\end{figure}

\model\ provides a flexible framework that can be applied to a diverse range of complex visual tasks. We evaluate \model\ on 4 tasks that require capabilities ranging from spatial reasoning, reasoning about multiple images, knowledge retrieval, and image generation and manipulation. Fig.~\ref{fig:task_io_modules} summarizes the inputs, outputs, and modules used for these tasks. We now describe these tasks, their evaluation settings, and the choice of in-context examples. 

\subsection{Compositional Visual Question Answering}\label{sec:gqa_task}
\model\ is compositional by construction which makes it suitable for the compositional, multi-step visual question answering task: \gqa\ \cite{Hudson2019GQA}. Modules for the GQA task include those for open vocabulary localization, a VQA module, functions for cropping image regions given bounding box co-ordinates or spatial prepositions (such as  \emph{above}, \emph{left}, \etc), module to count boxes, and 
a module to evaluate Python expressions. For example, consider the question: ``Is the small truck to the left or to the right of the people that are wearing helmets?". \model\ first localizes ``people wearing helmets", crops the region to the left (or right) of these people, checks if there is a ``small truck" on that side, and return ``left" if so and ``right" otherwise. \model\ uses the question answering module based on \vilt\ \cite{Kim2021ViLT}, but instead of simply passing the complex original question to \vilt, \model\ invokes it for simpler tasks like identifying the contents within an image patch. As a result, our resulting \model\ for GQA is not only more interpretable than \vilt\ but also more accurate (Tab.~\ref{tab:gqa_test}). Alternatively, one could completely eliminate the need for a QA model like ViLT and use other systems like CLIP and object detectors, but we leave that for future investigation. 

\noindent\textbf{Evaluation.} In order to limit the money spent on generating programs with \gpt, we create a subset of \gqa\ for evaluation. Each question in \gqa\ is annotated with a question type. To evaluate on a diverse set of question types ($\sim100$ detailed types), we randomly sample up to $k$ samples per question type from the balanced \textit{val} ($k=5$) and \textit{testdev} ($k=20$) sets. \\ 

\begin{figure*}[t]
  \centering
  \includegraphics[width=\linewidth]{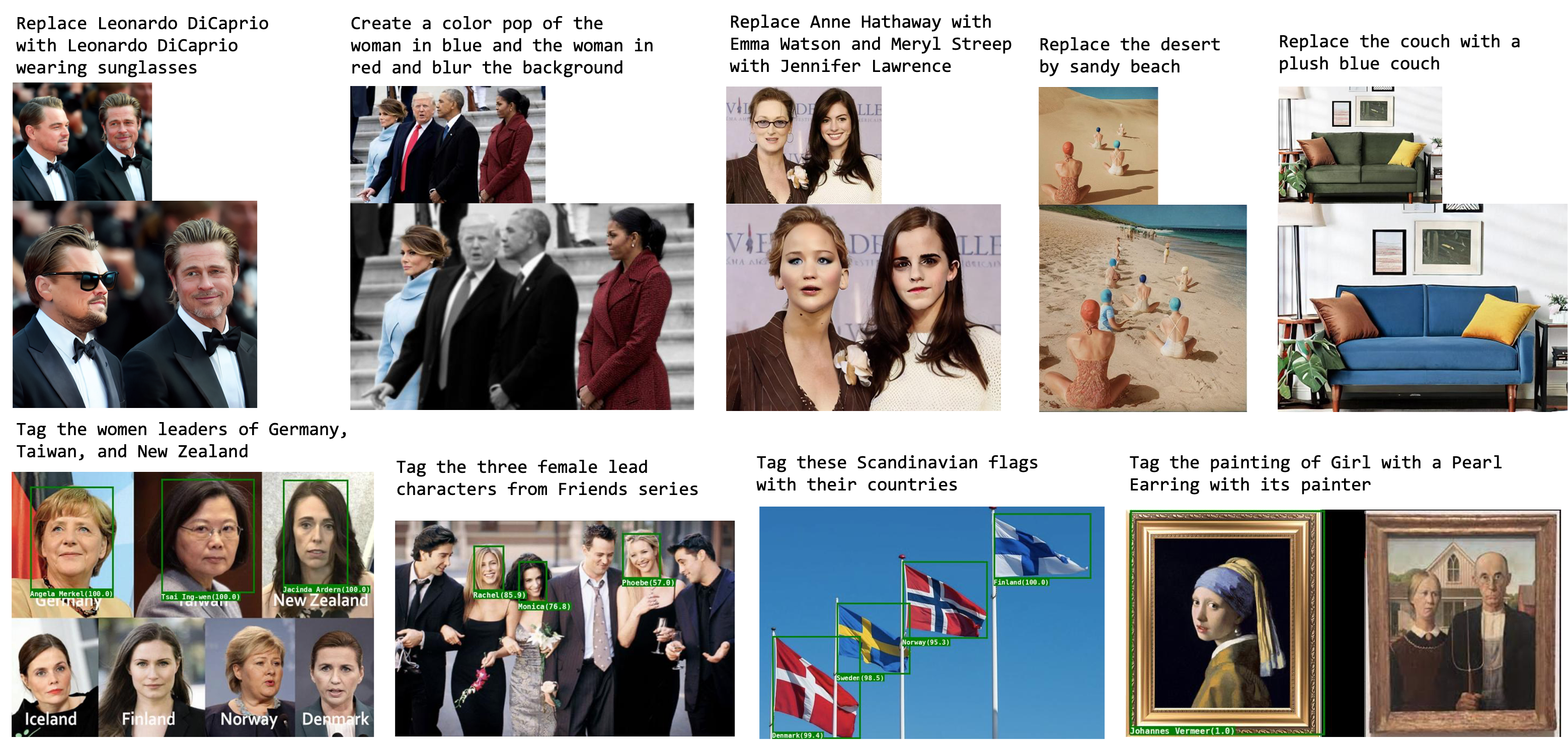}
   \caption{Qualitative results for image editing (top) and knowledge tagging tasks (bottom).}
   \label{fig:qualitative}
\end{figure*}

\noindent\textbf{Prompts.} We manually annotate $31$ random questions from the balanced \textit{train} set with desired \model\ programs. Annotating questions with programs is easy and requires writing down the chain of reasoning required to answer that particular question. We provide a smaller subset of in context examples to \gpt, randomly sampled from this list to reduce the cost of answering each GQA question.

\subsection{Zero-Shot Reasoning on Image Pairs}

VQA models are trained to answer questions about a single image. In practice, one might require a system to answer questions about a collection of images. For example, a user may ask a system to parse their vacation photo album and answer the question: ``Which landmark did we visit, the day after we saw the Eiffel Tower?''. Instead of assembling an expensive dataset and training a multi-image model, we demonstrate the ability of \model\ to use a single-image VQA system to solve a task involving multiple images without training on multi-image examples. 

We showcase this ability on the \nlvr\ \cite{Suhr2019nlvr2} benchmark, which involves verifying statements about image pairs. Typically, tackling the \nlvr\ challenge requires training custom architectures that take image pairs as input on \nlvr's train set. Instead, \model\ achieves this by decomposing a complex statement into simpler questions about individual images and a python expression involving arithmetic and logical operators and answers to the image-level questions. The VQA model \viltvqa\ is used to get image-level answers, and the python expression is evaluated to verify the statement.

\noindent\textbf{Evaluation.} We create a small validation set by sampling $250$ random samples from the \nlvr\ \textit{dev} set to guide prompt selection, and test generalization on \nlvr's full public \textit{test} set.

\noindent\textbf{Prompts.} We sample and annotate \model\ programs for $16$ random statements in the \nlvr\ \textit{train} set. Since some of these examples are redundant (similar program structure) we also create a curated subset of $12$ examples by removing $4$ redundant ones.

\subsection{Factual Knowledge Object Tagging}\label{sec:knowtag}
We often want to identify people and objects in images whose names are unknown to us. For instance, we might want to identify celebrities, politicians, characters in TV shows, flags of countries, logos of corporations, popular cars and their manufacturers, species of organisms, and so on. Solving this task requires not only localizing people, faces, and objects but also looking up factual knowledge in an external knowledge base to construct a set of categories for classification, such as names of the characters on a TV show. We refer to this task as Factual Knowledge Object Tagging or Knowledge Tagging for short.

For solving Knowledge Tagging, \model\ uses \gpt\ as an implicit knowledge base that can be queried with natural language prompts such as ``List the main characters on the TV show Big Bang Theory separated by commas." This generated category list can then be used by a \clip\ image classification module that classifies image regions produced by localization and face detection modules. 
\model's program generator automatically determines whether to use a face detector or an open-vocabulary localizer depending on the context in the natural language instruction. \model\ also estimates the maximum size of the category list retrieved. For instance, ``Tag the logos of the top 5 german car companies" generates a list of 5 categories, while ``Tag the logos of german car companies" produces a list of arbitrary length determined by \gpt\ with a cut-off at 20. This allows users to easily control the noise in the classification process by tweaking their instructions.

\noindent\textbf{Evaluation.} To evaluate \model\ on this task, we annotate 100 tagging instructions across 46 images that require external knowledge to tag 253 object instances including personalities across pop culture, politics, sports, and art, as well as a  varieties of objects (\eg cars, flags, fruits, appliances, furniture \etc). For each instruction, we measure both localization and tagging performance via precision (fraction of predicted boxes that are correct) and recall (fraction of ground truth objects that are correctly predicted). Tagging metrics require both the predicted bounding box and the associated tag or class label to be correct, while localization ignores the tag.  To determine localization correctness, we use an IoU threshold of 0.5. We summarize localization and tagging performance by F1 scores (harmonic mean of the average precision and recall across instructions).

\noindent\textbf{Prompts.} We create 14 in-context examples for this task. Note that the instructions for these examples were hallucinated \ie no images were associated with these examples.

\subsection{Image Editing with Natural Language}\label{sec:imgedit}
Text to image generation has made impressive strides over the last few years with models like DALL-E~\cite{Ramesh2021dalle}, Parti~\cite{Yu2022parti}, and Stable Diffusion~\cite{Rombach2022stableDiffusion}. However, it is still beyond the capability of these models to handle prompts like "Hide the face of Daniel Craig with :p" (\textbf{de-identification} or \textbf{privacy preservation}), or "Create a color pop of Daniel Craig and blur the background" (\textbf{object highlighting}) even though these are relatively simple to achieve programmatically using a combination of face detection, segmentation and image processing modules. Achieving a sophisticated edit such as "Replace Barack Obama with Barack Obama wearing sunglasses" (\textbf{object replacement}), first requires identifying the object of interest, generating a mask of the object to be replaced and then invoking an image inpainting model (we use Stable Diffusion) with the original image, mask specifying the pixels to replace, and a description of the new pixels to generate at that location. \model, when equipped with the necessary modules and example programs, can handle very complex instructions with ease.

\noindent\textbf{Evaluation.} To test \model\ on the image editing instructions for de-identification, object highlighting, and object replacement, we collect 107 instructions across 65 images. We manually score the predictions for correctness and report accuracy. Note that we do not penalize visual artifacts for the object replacement sub-task which uses Stable Diffusion as long as the generated image is semantically correct.

\noindent\textbf{Prompts.} Similar to knowledge tagging, we create 10 in-context examples for this task with no associated images. 


\section{Experiments and Analysis}
Our experiments evaluate the effect of number of prompts on GQA and NLVR performance (Sec.~\ref{sec:prompt_size}), generalization of \model\ on the four tasks comparing various prompting strategies (Sec.~\ref{sec:generalization}), analyze the sources of error for each task (Fig.~\ref{fig:errors}), and study the utility of visual rationales for diagnosing errors and improving \model's performance through instruction tuning (Sec.~\ref{sec:utility}).



\subsection{Effect of prompt size}\label{sec:prompt_size}
\begin{figure}[t]
  \centering
  \includegraphics[width=1\linewidth]{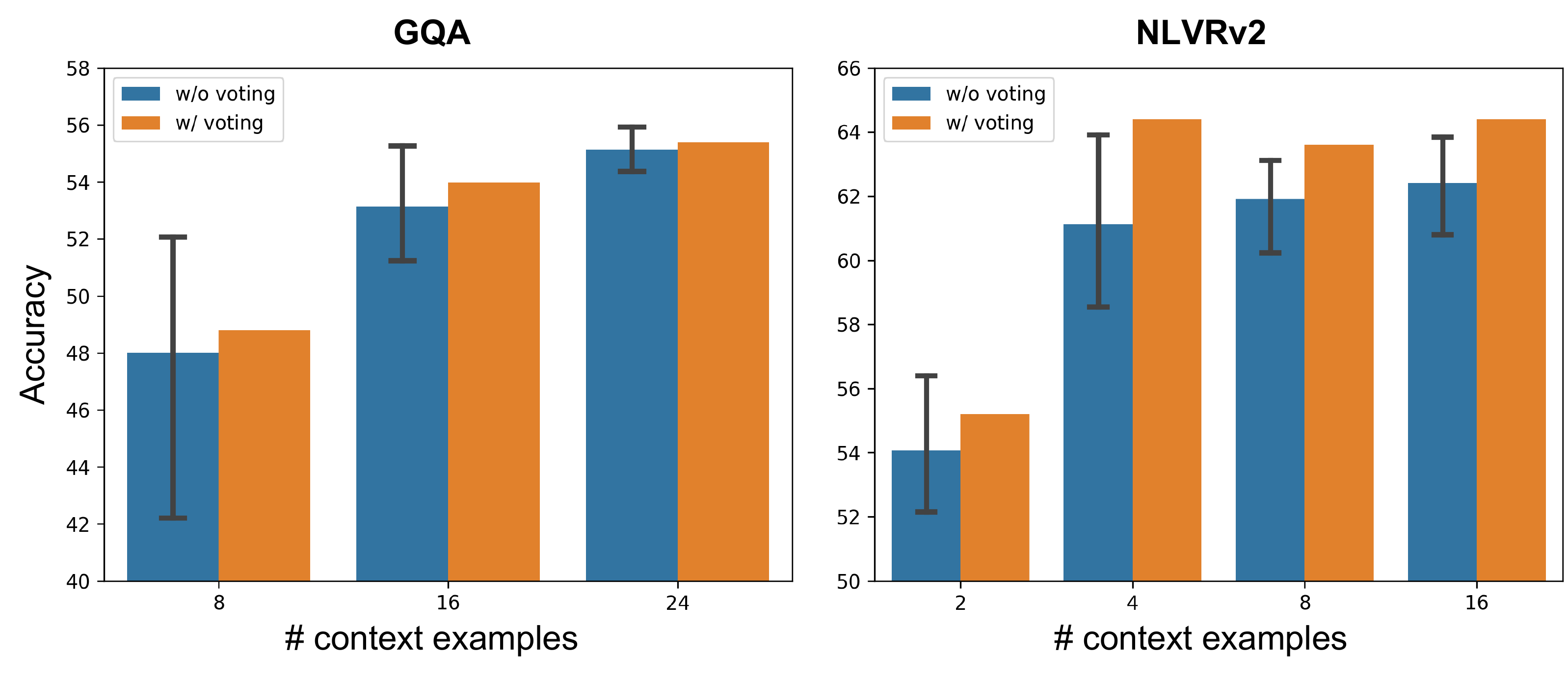}
   \caption{\textbf{Performance improves with number of in-context examples on \gqa\ and \nlvr\ validation sets.} The error bars represent $95\%$ confidence interval across $5$ runs. Predictions from the same runs are used for majority voting. (Sec.~\ref{sec:prompt_size})}
   \label{fig:effect_of_num_prompts}
\end{figure}
Fig.~\ref{fig:effect_of_num_prompts} shows that validation performance increases progressively with the number of in-context examples used in the prompts for both GQA and NLVR. Each run randomly selects a subset of the annotated in-context examples based on a random seed. We also find that majority voting across the random seeds leads to consistently better performance than the average performance across runs. This is consistent with findings in Chain-of-Thought~\cite{Wei2022CoT} reasoning literature for math reasoning problems~\cite{Wang2022SelfConsistencyCoT}. On NLVR, the performance of \model\ saturates with fewer prompts than \gqa. We believe this is because \nlvr\ programs require fewer modules and hence fewer demonstrations for using those modules than \gqa.

\subsection{Generalization}\label{sec:generalization} 

\begin{table}
  \centering
  \resizebox{0.9\columnwidth}{!}{\begin{tabular}{@{}lcccc@{}}
    \toprule
    Method & \makecell{Prompting\\strategy} & Runs & \makecell{Context examples \\ per run} & Accuracy \\
    \midrule
    \viltvqa\ & - & 1 & - & 47.8 \\ \hline
    \model\ & curated & 1 & 20 & 50.0 \\
    \model\ & random & 1 & 24 & 48.2 \\
    \model\ & voting & 5 & 24 & \textbf{50.5} \\
    \bottomrule
  \end{tabular}}
  \caption{\textbf{\gqa\ \textit{testdev} results.} We report performance on a subset of the original GQA \textit{testdev} set as described in Sec.~\ref{sec:gqa_task}.}
  \label{tab:gqa_test}
\end{table}
\begin{table}[t]
  \centering
  \resizebox{1\columnwidth}{!}{\begin{tabular}{@{}lccccc@{}}
    \toprule
    Method & \makecell{Prompting\\strategy} & Finetuned & Runs & \makecell{Context examples\\per run} & Accuracy \\
    \midrule
    \viltnlvr & - & \cmark & 1 & - & 76.3 \\ \hline
    \model\ & curated & \xmark & 1 & 12 & 61.8 \\
    \model & random & \xmark & 1 & 16 & 61.3 \\
    \model\ & voting & \xmark & 5 & 16 & \textbf{62.4} \\
    \bottomrule
  \end{tabular}}
  \caption{\textbf{\nlvr\ \textit{test} results.} \model\ performs NLVR zero-shot \ie without training any module on image pairs. \viltnlvr, a \vilt\ model finetuned on \nlvr, serves as an upper bound.}
  \label{tab:nlvr_test}
\end{table}

\begin{table}
  \centering
  \resizebox{1\columnwidth}{!}{\begin{tabular}{lccccccc}
  \toprule
\multirow{2}{*}{Instructions} & \multicolumn{3}{c}{Tagging} & & \multicolumn{3}{c}{Localization} \\ \cline{2-4} \cline{6-8} 
         & precision  & recall  & F1   & & precision    & recall   & F1     \\ \midrule
Original             & 69.0         & 59.1    & 63.7 & & 87.2         & 74.9     & 80.6   \\
Modified             & 77.6       & 73.9    & 75.7 & & 87.4         & 82.5     & 84.9  \\ \bottomrule
\end{tabular}}
  \caption{\textbf{Knowledge tagging results.} The table shows performance on original instructions as well as modified instructions created after inspecting visual rationales to understand instance-specific sources of errors.}
  \label{tab:knowtag}
\end{table}
\begin{table}[t]
  \centering
  \resizebox{0.45\columnwidth}{!}{\begin{tabular}{@{}lcc@{}}
    \toprule
    & Original & Modified \\
    \midrule
    Accuracy & 59.8 & 66.4 \\
    \bottomrule
  \end{tabular}}
  \caption{\textbf{Image editing results.} We manually evaluate each prediction for semantic correctness.}
  \label{tab:imgedit}
\end{table}


\noindent\textbf{GQA.} In Tab.~\ref{tab:gqa_test} we evaluate different prompting strategies on the GQA \textit{testdev} set. For the largest prompt size evaluated on the \textit{val} set ($24$ in-context examples), we compare the random strategy consisting of the \model's best prompt chosen amongst 5 runs on the validation set (each run randomly samples in-context examples from 31 annotated examples) and the majority voting strategy which takes maximum consensus predictions for each question across 5 runs. While ``random" prompts only slightly outperform \viltvqa, voting leads to a significant gain of $2.7$ points. This is because voting across multiple runs, each with a different set of in-context examples, effectively increases the total number of in-context examples seen for each prediction. We also evaluate a manually curated prompt consisting of $20$ examples - $16$ from the $31$ annotated examples, and $4$ additional hallucinated examples meant to provide a better coverage for failure cases observed in the validation set. The curated prompt performs just as well as the voting strategy while using $5\times$ less compute, highlighting the promise of prompt engineering. \\

\noindent\textbf{NLVR.} Tab.~\ref{tab:nlvr_test} shows performance of \model\ on the \nlvr\ \textit{test} set and compares random, voting, and curated prompting strategies as done with GQA. While \model\ performs the NLVR task \textit{zero-shot} without ever training on image pairs, we report \viltnlvr, a \vilt\ model finetuned on \nlvr\, as an upper bound on performance. While several points behind the upper bound, \model\ shows strong zero-shot performance using only a single-image VQA model for image understanding, and an LLM for reasoning. Note that, \model\ uses \viltvqa\ for its VQA module which is trained on \vqavtwo\, a single image question answer task, but not \nlvr. \\

\noindent\textbf{Knowledge Tagging.} Tab.~\ref{tab:knowtag} shows localization and tagging performance for the Knowledge Tagging task. All instructions for this task not only require open vocabulary localization but also querying a knowledge base to fetch the categories to tag localized objects with. This makes it an impossible task for object detectors alone. With the original instructions, \model\ achieves an impressive $63.7\%$ F1 score for tagging, which involves both correctly localizing and naming the objects, and $80.6\%$ F1 score for localization alone. Visual rationales in \model\ allow further performance gains by modifying the instructions. See Fig.~\ref{fig:qualitative} for qualitative examples and Sec.~\ref{sec:utility} for more details on \textit{instruction tuning}.  \\

\noindent\textbf{Image Editing.} Tab.~\ref{tab:imgedit} shows the performance on the language-guided image editing task. Fig.~\ref{fig:qualitative} shows the wide range of manipulations possible with the current set of modules in \model\ including face manipulations, highlighting one or more objects in the image via stylistic effects like color popping  and background blur, and changing scene context by replacing key elements in the scene (\eg desert).

\begin{figure}[t]
  \centering
  \fbox{\includegraphics[width=0.9\linewidth]{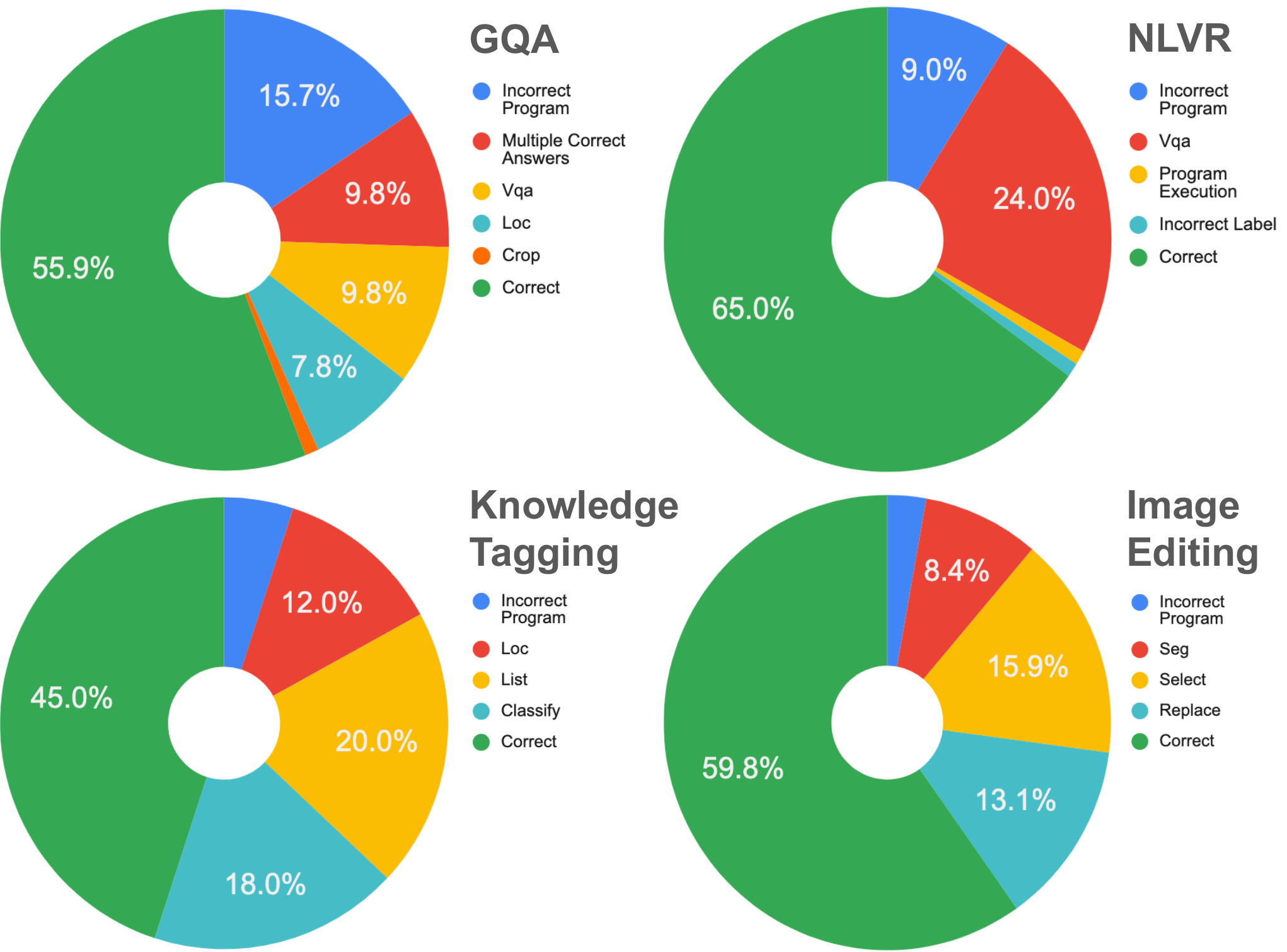}}
   \caption{\textbf{Sources of error in \model.}} 
   \label{fig:errors}
   \vspace{-1.0em}
\end{figure}
\begin{figure*}[t]
  \centering
  \includegraphics[width=\linewidth]{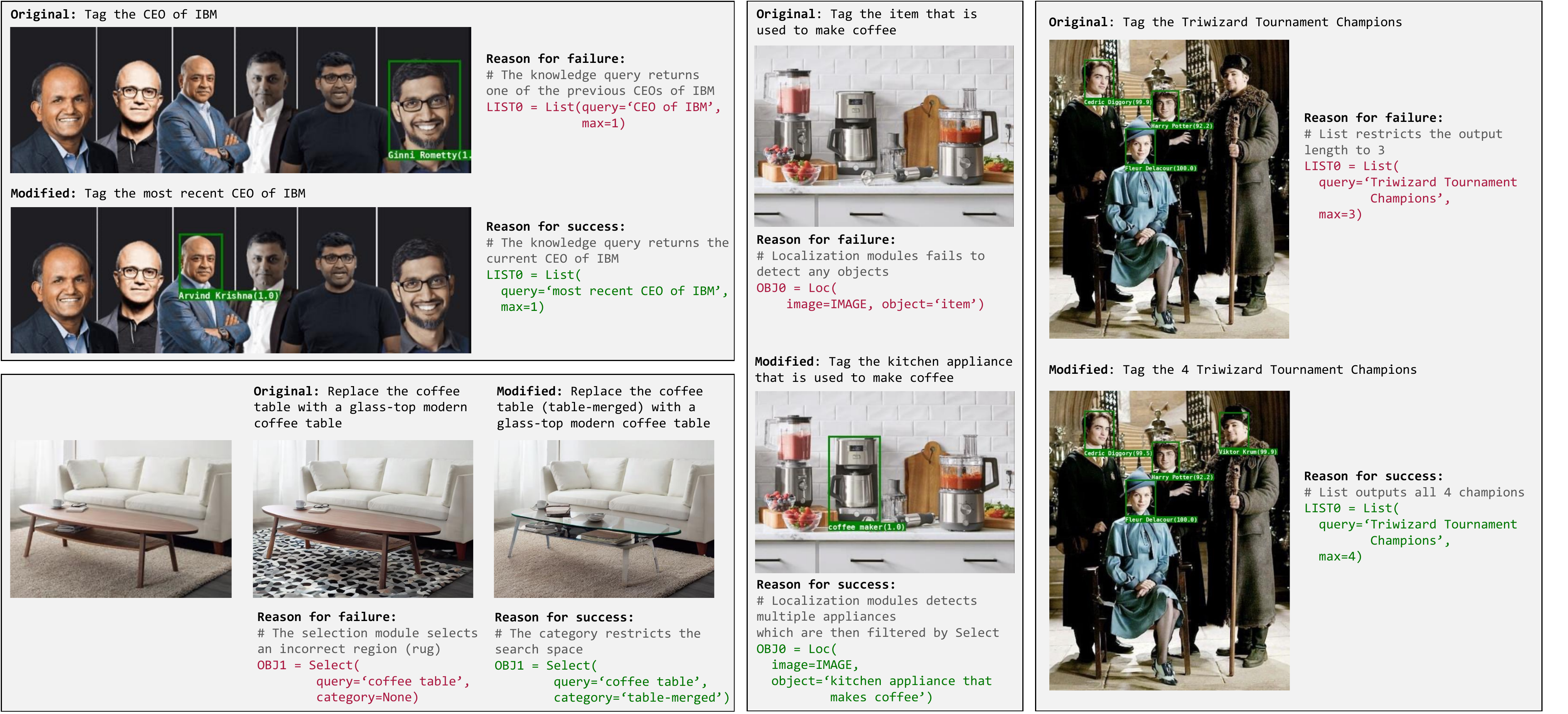}
   \caption{\textbf{Instruction tuning using visual rationales.} By revealing the reason for failure, \model\ allows a user to modify the original instruction to improve performance.}
   \vspace{-1em}
   \label{fig:user_feedback}
\end{figure*}
\subsection{Utility of Visual Rationales}\label{sec:utility}
\noindent\textbf{Error Analysis.}
Rationales generated by \model\ allow a thorough analysis of failure modes as shown in Fig.~\ref{fig:errors}. For each task, we manually inspect rationales for $\sim100$ samples to break down the sources of errors. Such analysis provides a clear path towards improving performance of \model\ on various tasks. For instance, since incorrect programs are the leading source of errors on GQA affecting $16\%$ of samples, performance on GQA may be improved by providing more in-context examples similar to the instructions that \model\ currently fails on. 
Performance may also be improved by upgrading models used to implement the high-error modules to more performant ones. For example, replacing the \viltvqa\ model with a better VQA model for NLVR could improve performance by up to $24\%$. Similarly, improving models used to implement ``List" and ``Select" modules, the major sources of error for knowledge tagging and image editing tasks, could significantly reduce errors. \\




\noindent\textbf{Instruction tuning.} To be useful, a visual rationale must ultimately allow users to improve the performance of the system on their task. For knowledge tagging and image editing tasks, we study if visual rationales can help a user modify or \textit{tune} the instructions to achieve better performance. Fig.~\ref{fig:user_feedback} shows that modified instructions: (i) result in a better query for the localization module (\eg ``kitchen appliance" instead of ``item"); (ii) provide a more informative query for knowledge retrieval (\eg ``most recent CEO of IBM" instead of ``CEO of IBM"); (iii) provide a category name (\eg ``table-merged") for the Select module to restrict search to segmented regions belonging to the specified category; or (iv) control the number of classification categories for knowledge tagging through the \lstinline{max} argument in the List module. 
Tables~\ref{tab:knowtag} and ~\ref{tab:imgedit} show that instruction tuning results in significant gains for knowledge tagging and image editing tasks.
\section{Conclusion}
\model\ proposes visual programming as a simple and effective way of bringing the reasoning capabilities of LLMs to bear on complex visual tasks. \model\ demonstrates strong performance while generating highly interpretable visual rationales. Investigating better prompting strategies and exploring new ways of incorporating user feedback to improve the performance of neuro-symbolic systems such as \model\ is an exciting direction for building the next generation of general-purpose vision systems.

\section{Acknowledgement}
We thank Kanchan Aggarwal for helping with the annotation process for the image editing and knowledge tagging tasks. We are also grateful to the amazing Hugging Face ecosystem for simplifying the use of state-of-the-art neural models for implementing \model\ modules. 

{\small
\bibliographystyle{ieee_fullname}
\bibliography{egbib}
}

\appendix
\clearpage
\section{Appendix}
This appendix includes
\begin{itemize}
    \vspace{-0.5em} \item Task prompts for \model (Sec.~\ref{sec:app_task_prompt})
    \vspace{-0.5em} \item Module implementation details (Sec.~\ref{sec:app_module_details})
    \vspace{-0.5em} \item Many more qualitative results with visual rationales for both successful and failure cases can be found at \href{https://prior.allenai.org/projects/visprog}{https://prior.allenai.org/projects/visprog}.
\end{itemize}

\subsection{Task Prompts}\label{sec:app_task_prompt}
\begin{figure}[h!]
    \centering
    \frame{\includegraphics[width=1.0\linewidth,trim={0 0.5em 0 0},clip]{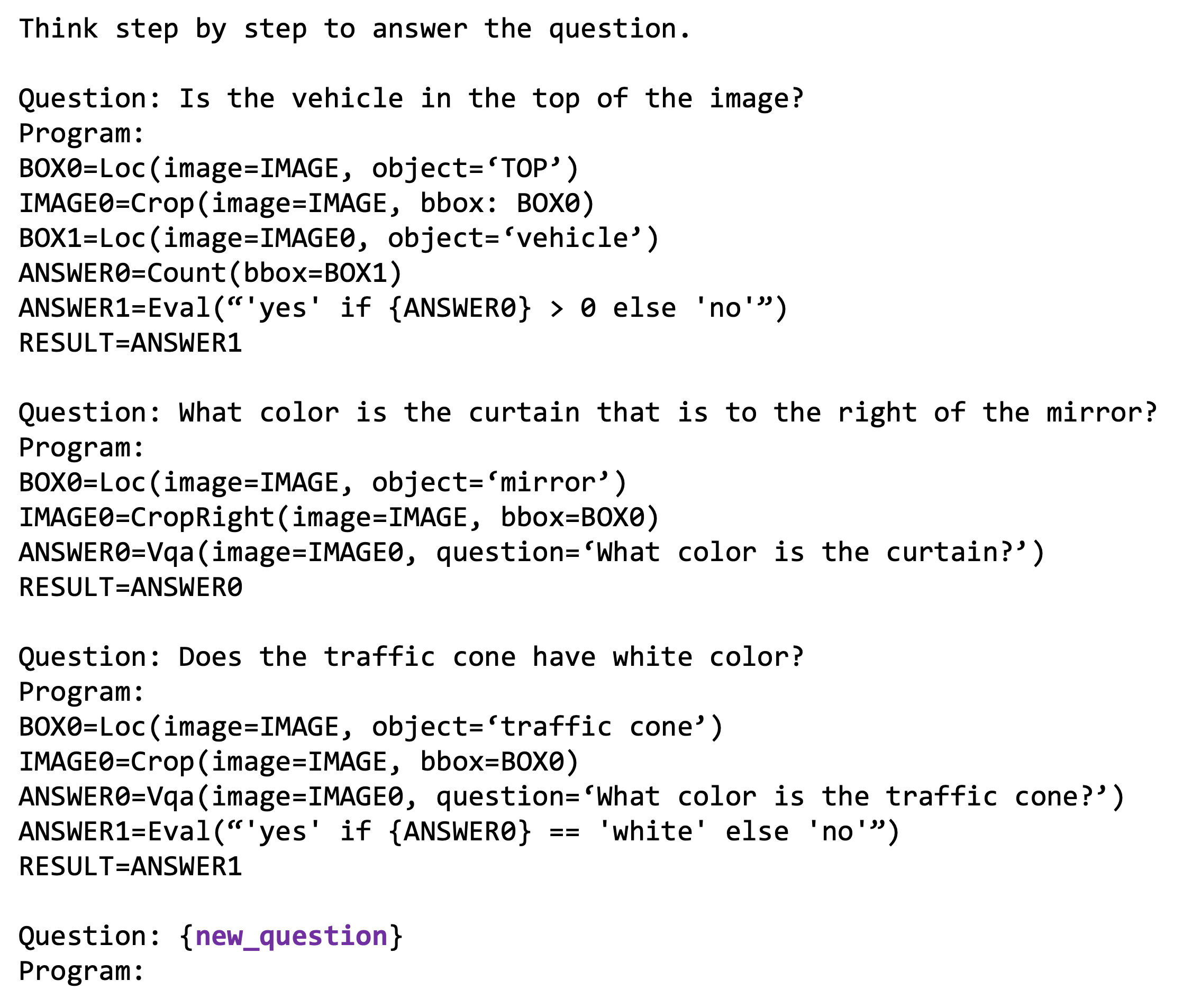}}
    \caption{\textbf{GQA prompt}}
    \label{fig:gqa_prompt}
\end{figure}

\begin{figure}
    \centering
    \frame{\includegraphics[width=1.0\linewidth,trim={0 1em 0 0},clip]{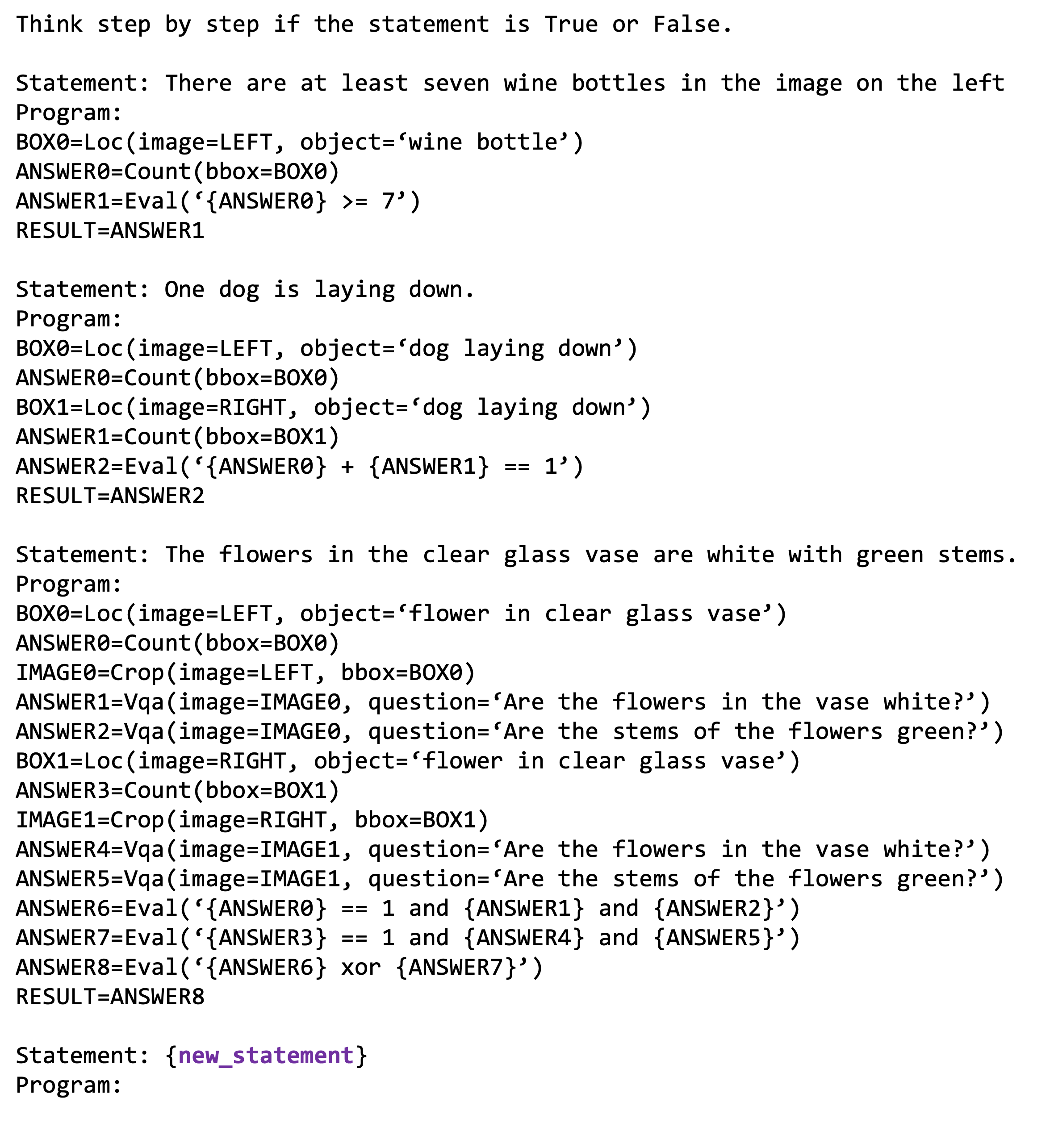}}
    \caption{\textbf{NLVR prompt}}
    \label{fig:nlvr_prompt}
\end{figure}

\begin{figure}[t]
    \centering
    \frame{\includegraphics[width=1\linewidth]{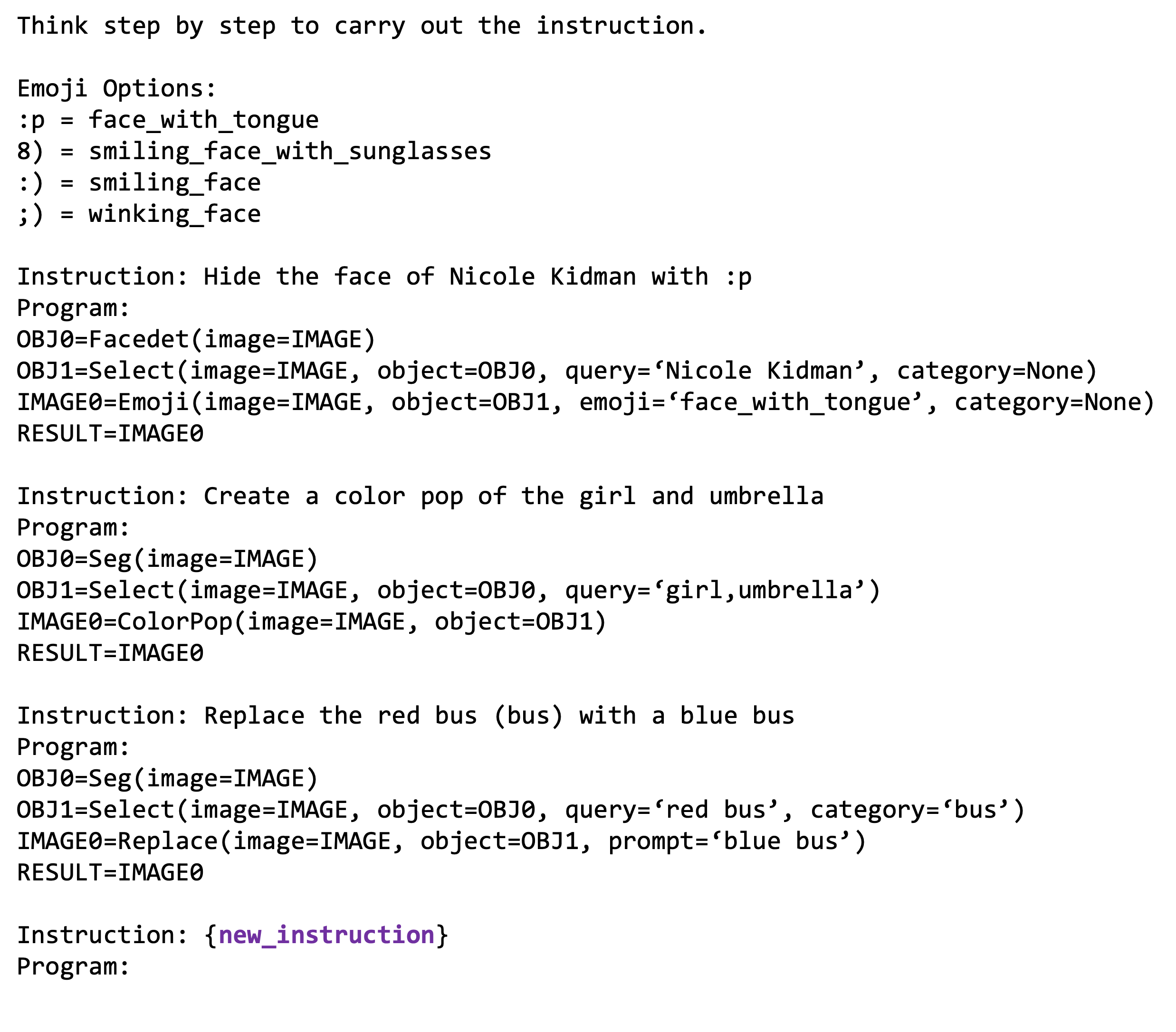}}
    \caption{\textbf{Image editing prompt}. Note that the prompt includes a mapping of emojis to their names in the AugLy~\cite{Papakipos2022AugLyDA} library that is used to implement \lstinline{Emoji} module. The third example shows how to provide the \lstinline{category} value for the \lstinline{Select} module.}
    \label{fig:imgedit_prompt}
\end{figure}

\begin{figure}[t]
    \centering
    \frame{\includegraphics[width=1\linewidth,trim={0 1em 0 0},clip]{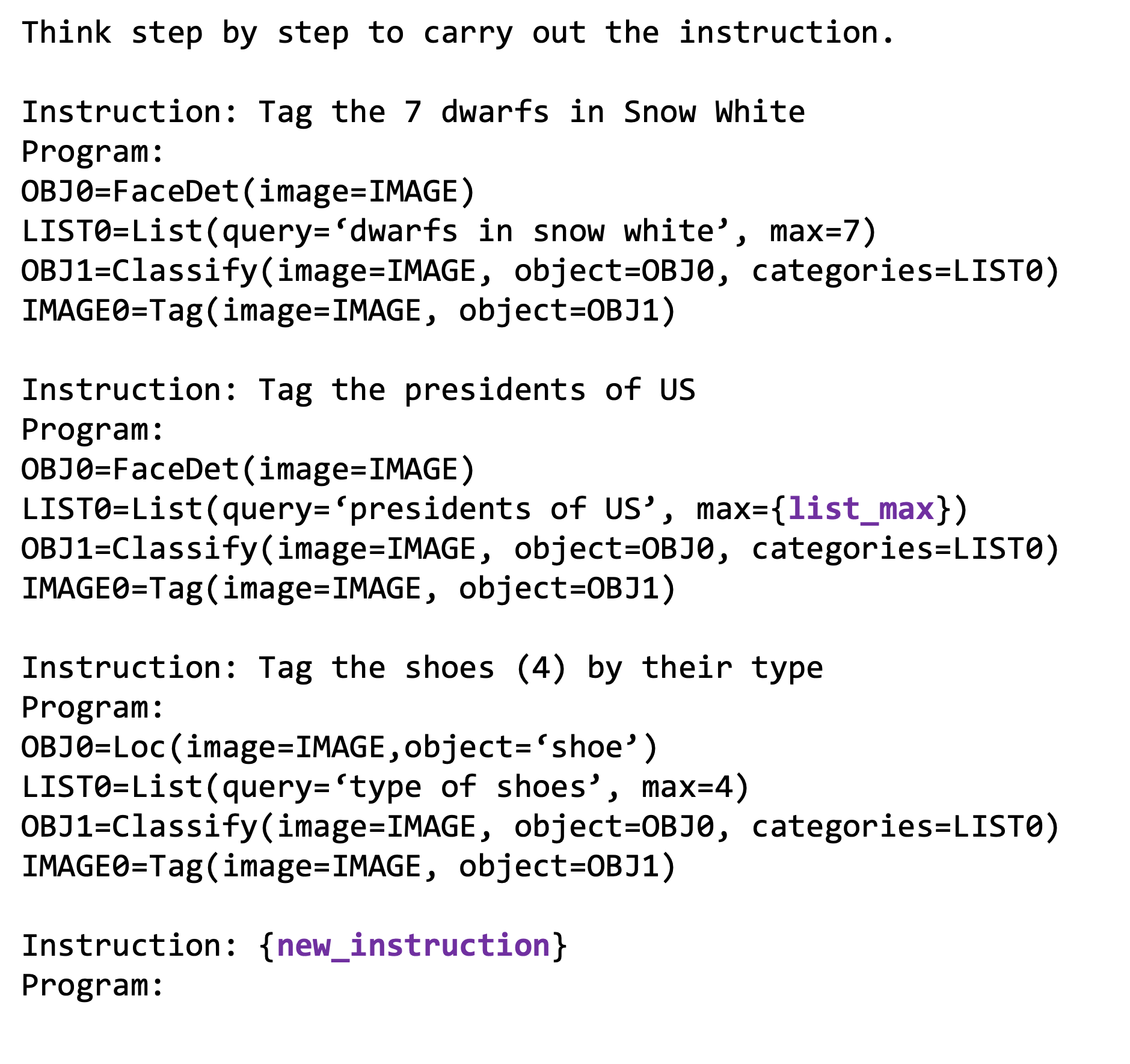}}
    \caption{\textbf{Knowledge tagging prompt.} Note that the prompt has an additional placeholder to configure the default \lstinline{max} value for \lstinline{List} module. While the first example infers \lstinline{max} from a natural instruction, the third example demonstrates how a user might minimally augment a natural instruction to provide argument values.}
    \label{fig:tag_prompt}
\end{figure}

We show the prompt structures for GQA (Figure~\ref{fig:gqa_prompt}), NLVR (Figure~\ref{fig:nlvr_prompt}), knowledge tagging (Figure~\ref{fig:tag_prompt}), and language-guided image editing (Figure~\ref{fig:imgedit_prompt}) tasks with 3 in-context examples each. 

\subsection{Module Details}\label{sec:app_module_details}
To help understand the generated programs better, we now provide a few implementation details about some of the modules. \\

\noindent\textbf{Select.} The module takes a \lstinline{query} and a \lstinline{category} argument. When the \lstinline{category} is provided, the selection is only performed over the regions that have been identified as belonging to that category by a previous module in the program (typically the \lstinline{Seg} module). If \lstinline{category} is \lstinline{None}, the selection is performed over all regions. The query is the text to be used for region-text scoring to perform the selection. We use CLIP-ViT~\cite{clip} to select the region with the maximum score for the query. When the query contains multiple phrases separated by commas, the highest-scoring region is selected for each phrase. \\

\noindent\textbf{Classify.} The \lstinline{Classify} module takes lists of object regions and categories and tries to assign one of the categories to each region. For simplicity, we assume the images in the tagging task has at most 1 instance of each category. The \lstinline{Classify} module operates differently based on whether the category list has 1 or more elements. If the category list has only 1 element, the category is assigned to the region with the highest CLIP score, similar to the \lstinline{Select} module. When more than one category is provided, first, each region is assigned the category with the best score. Due to classification errors, this can lead to multiple regions being assigned the same category. Therefore, for each of the assigned categories (excluding the ones that were not assigned to any region), we perform a de-duplication step that retains only the maximum scoring region for each category. \\

\begin{figure}[h]
    \centering
    \frame{\includegraphics[width=1.0\linewidth]{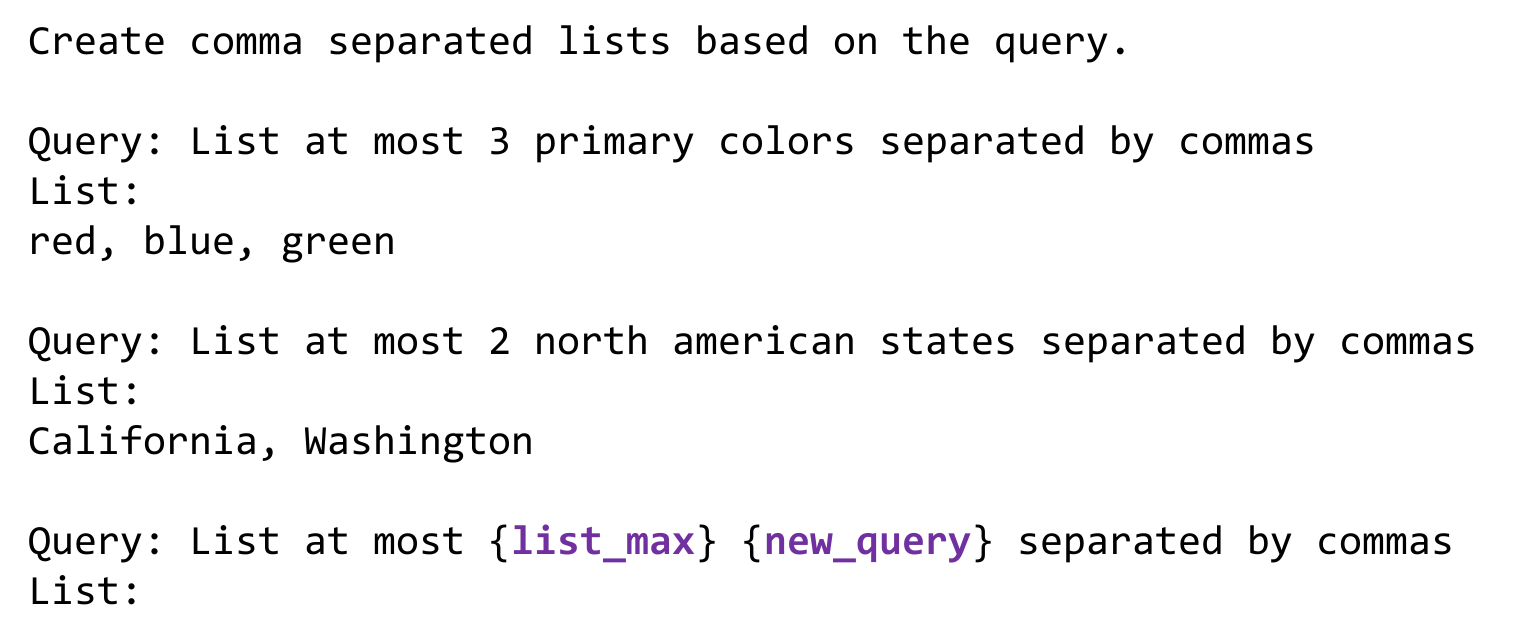}}
    \caption{\textbf{Prompt for the \lstinline{List} module.} \lstinline{list_max} denotes the default maximum list length and \lstinline{new_query} is the placeholder for the new retrieval query}
    \label{fig:list_prompt}
\end{figure}

\noindent\textbf{List.} The \lstinline{List} module uses GPT3 to create a flexible and powerful knowledge retriever. Fig.~\ref{fig:list_prompt} shows the prompt provided to GPT3 to retrieve factual knowledge.

\end{document}